# INTERVAL STRUCTURE: A Framework for Representing Uncertain Information

**S.K.M. Wong, L.S. Wang and Y.Y. Yao**
Department of Computer Science
University of Regina
Regina, Saskatchewan
Canada S4S 0A2

## Abstract

In this paper, a unified framework for representing uncertain information based on the notion of an interval structure is proposed. It is shown that the lower and upper approximations of the rough-set model, the lower and upper bounds of incidence calculus, and the belief and plausibility functions all obey the axioms of an interval structure. An interval structure can be used to synthesize the decision rules provided by the experts. An efficient algorithm to find the desirable set of rules is developed from a set of sound and complete inference axioms.

## 1 INTRODUCTION

In decision making, we often find ourselves in a state of uncertainty. This might stem from either a lack of knowledge, or from the incompleteness or unreliability of the information at our disposal. To make decisions under such circumstances, it is crucial to choose an appropriate structure to represent uncertain information.

Although probability theory is the standard method for dealing with uncertainty, other constructs such as rough sets, fuzzy sets, and belief functions play an important role in the design of expert systems. In these non-standard methods, uncertainty is represented by an *interval* within which the truth lies. Pawlak (1982, 1984) introduced the concept of rough sets, which characterizes an ordinary set by a *lower* and an *upper approximation*. The lower approximation contains the objects definitely belonging to the set, whereas the upper approximation contains the objects possibly belonging to the set. In the study of incidence calculus, Bundy (1985, 1986) examined the lower and upper *bounds* of incidences of a set of propositions. Lower bounds represent situations in which the propositions are definitely true, and upper bounds represent situations in which the propositions could be true. In the theory of fuzzy sets, the *core* (a lower approximation) of a fuzzy set is defined by collecting all elements with complete membership, while the *support* (an upper approximation) is defined by collecting all elements with non-zero membership (Zadeh, 1965; Dubois and Prade, 1990). Recently, Yao and Wong (1991) studied the rough-set and fuzzy-set models within the Bayesian decision theoretic framework. In this approach, a set may be approximated by different levels of lower and upper bounds depending on the application. It should be noted that all the bounds considered in these models are non-numeric bounds (crisp sets); bounds expressed in terms of non-crisp sets were studied by Dubois and Prade (1990).

The numeric belief and plausibility functions proposed by Shafer (1976) can be interpreted as the lower and upper bounds of probability functions (Dempster, 1967; Dubois and Prade, 1985; Halpern and Fagin, 1990). These numeric bounds are in fact closely related to non-numeric bounds. The basic idea of rough sets was implicitly used by Shafer (1976) in defining the notions of coarsening and refinement of a frame. More recently, Grzymala-Busse (1987), Wong and Lingras (1989) investigated the relationships between belief/plausibility functions and lower/upper approximations in the rough-set model in an attempt to establish a linkage between numeric and non-numeric representations of uncertain information.

The results of the studies mentioned above seem to suggest that there exists a common framework for modeling uncertainty. This paper introduces the notion of *interval structure* to represent uncertain information. Both non-numeric and numeric bounds will be analyzed in this framework. We will show that the lower and upper approximations of the rough-set model, the lower and upper bounds in incidence calculus, and the belief and plausibility functions all obey the axioms of an interval structure. To demonstrate the usefulness of such a structure, we apply the techniques developed here to synthesize the knowledge provided by the experts. The process of such synthesis not only provides a desirable set of decision rules, but also clearly demonstrates the explicit structure of these rules.

## 2 INTERVAL STRUCTURE INDUCED BY A COMPATIBILITY RELATION

Let $W = \{w_1, w_2, \ldots, w_m\}$ and $\Theta = \{\theta_1, \theta_2, \ldots, \theta_n\}$ represent two finite universes of interest. The relationships between the elements of $W$ and $\Theta$ can be characterized by a *compatibility* relation (Shafer, 1986). A compatibility relation is defined as a subset of pairs $(w, \theta)$ in the Cartesian product $W \times \Theta$. An element $w \in W$ is *compatible* with an element $\theta \in \Theta$, written $w\ C\ \theta$, if the $w$ is related to $\theta$. In reality, the formulation and interpretation of $W$ and $\Theta$ and the compatibility relation between these two sets depend very much on the available knowledge and the domain of applications. For example, in a medical diagnosis system, $W$ can be a set of symptoms and $\Theta$ a set of diseases. A symptom $w$ is said to be compatible with a disease $\theta$ if any patient with symptom $w$ may have contracted the disease $\theta$. Without loss of generality, we may assume that for any $w \in W$ there exists a $\theta \in \Theta$ with $w\ C\ \theta$, and vice versa.

A compatibility relation $C$ between $W$ and $\Theta$ can be equivalently defined by a multi-valued mapping, $\gamma : W \longrightarrow 2^\Theta$, as (Dempster, 1967; Shafer, 1986):

$$\gamma(w) = \{\theta \in \Theta \mid w\ C\ \theta\}. \quad (1)$$

Such a mapping $\gamma$ induces a function, $\Gamma : 2^W \longrightarrow 2^\Theta$, namely, for $X \in 2^W$,

$$\Gamma(X) = \bigcup_{w \in X} \gamma(w). \quad (2)$$

Note that function $\Gamma : 2^W \longrightarrow 2^\Theta$ is not necessarily an onto mapping, i.e., not every subset of $\Theta$ has a *preimage* in $2^W$. Therefore, it may not be possible to define the inverse of $\Gamma$ for every subset of $\Theta$. Nevertheless, we can define a lower inverse mapping $\underline{\Gamma^{-1}} : 2^\Theta \longrightarrow 2^W$, and an upper inverse mapping $\overline{\Gamma^{-1}} : 2^\Theta \longrightarrow 2^W$ as:

$$\underline{\Gamma^{-1}}(A) = \{w \in W \mid \gamma(w) \subseteq A\}, \quad (3)$$

and

$$\overline{\Gamma^{-1}}(A) = \{w \in W \mid \gamma(w) \cap A \neq \emptyset\}. \quad (4)$$

For an arbitrary subset $A \in 2^\Theta$, the set $\underline{\Gamma^{-1}}(A)$ consists of the elements in $W$ compatible with only those elements in $A$, while the set $\overline{\Gamma^{-1}}(A)$ consists of the elements in $W$ compatible with at least one element in $A$. In general, the lower and upper preimages are not necessarily the same. If information is transferred from subsets in $W$ to subsets in ●, or if subsets in $\Theta$ are characterized by subsets in $W$, the lower preimages $\underline{\Gamma^{-1}}(A)$ can be interpreted as the pessimistic estimation and the upper preimages $\overline{\Gamma^{-1}}(A)$ as the optimistic estimation of $A$. That is, the true preimage of $A$ lies in the interval $[\underline{\Gamma^{-1}}(A), \overline{\Gamma^{-1}}(A)]$.

We can extend this particular example to define the notion of an interval structure. Given two mappings $\underline{F} : 2^\Theta \longrightarrow 2^W$ and $\overline{F} : 2^\Theta \longrightarrow 2^W$, if $\underline{F}$ satisfies the axioms: for any subsets $A, B \in 2^\Theta$,

(L1) $\underline{F}(A \cup B) \supseteq \underline{F}(A) \cup \underline{F}(B)$,

(L2) $\underline{F}(A \cap B) = \underline{F}(A) \cap \underline{F}(B)$,

(L3) $\underline{F}(\emptyset) = \emptyset$,

(L4) $\underline{F}(\Theta) = W$,

$\overline{F}$ satisfies the axioms: for any subsets $A, B \in 2^\Theta$,

(U1) $\overline{F}(A \cup B) = \overline{F}(A) \cup \overline{F}(B)$,

(U2) $\overline{F}(A \cap B) \subseteq \overline{F}(A) \cap \overline{F}(B)$,

(U3) $\overline{F}(\emptyset) = \emptyset$,

(U4) $\overline{F}(\Theta) = W$,

and moreover, $\overline{F}(A) = W - \underline{F}(\neg A)$, where $\neg A = \Theta - A$ denotes the complement of $A$, the pair $F = (\underline{F}, \overline{F})$ is called an *interval structure*. For any subset $A \in 2^\Theta$, $\underline{F}(A)$ is called the lower bound and $\overline{F}(A)$ the upper bound of $A$ in $W$.

Given a lower bound mapping $\underline{F}$ satisfying axioms (L1)-(L4), the upper bound mapping can be easily obtained by the relationships, $\overline{F}(A) = W - \underline{F}(A)$, which automatically satisfies axioms (U1)-(U2). Likewise, given an upper bound mapping, one can obtain the corresponding lower bound mapping. Note that (L1)-(L4) are a set of independent axioms; (U1)-(U4) are also a set of independent axioms.

It can be easily seen that the interval $[\underline{\Gamma^{-1}}(A), \overline{\Gamma^{-1}}(A)]$ derived from a compatibility relation is an interval structure, i.e., $\underline{\Gamma^{-1}}$ satisfies axioms (L1)-(L4), $\overline{\Gamma^{-1}}$ satisfies axioms (U1)-(U4), and

$$\overline{\Gamma^{-1}}(A) = W - \underline{\Gamma^{-1}}(\neg A) = \neg\underline{\Gamma^{-1}}(\neg A). \quad (5)$$

It should be emphasized here that a compatibility relation provides only one of the possible ways to obtain an interval structure. In general, one can directly define an interval structure by demanding that axioms (L1)-(L4) and (U1)-(U4) are satisfied.

An interval structure also satisfies the following properties: for any $A, B \in 2^\Theta$,

(P1) $\underline{F}(A) \subseteq \overline{F}(A)$,

(P2) $\underline{F}(\neg A) = \neg\overline{F}(A)$, $\overline{F}(\neg A) = \neg\underline{F}(A)$,

(P3) $A \supseteq B \Longrightarrow (\underline{F}(A) \supseteq \underline{F}(B), \overline{F}(A) \supseteq \overline{F}(B))$.

We can equivalently define an interval structure $F$ by a *basic set assignment*, $j_F : 2^\Theta \to 2^W$, which satisfies the following axioms: for any $A, B \in 2^\Theta$,

(A1) $j_F(\emptyset) = \emptyset$,

(A2) $\bigcup_{A \subseteq \Theta} j_F(A) = W$,

(A3) $A \neq B \Longrightarrow (j_F(A) \cap j_F(B) = \emptyset)$.

Based on $j_F$, the lower and upper bounds of $A$ can be expressed as:

$$\underline{F}(A) = \bigcup_{B \subseteq A} j_F(B), \quad \overline{F}(A) = \bigcup_{A \cap B \neq \emptyset} j_F(B), \quad (6)$$

Conversely, from an interval structure $F$, one can construct the basic set assignment $j_F$ by:

$$j_F(A) = \underline{F}(A) - (\bigcup_{B \subset A} \underline{F}(B)). \quad (7)$$

A subset $A \in 2^\Theta$ with $j_F(A) \neq \emptyset$ is called a *focal set*. In the special case where the interval structure is induced by a compatibility relation, the basic set assignment can be expressed as:

$$j_F(A) = \{w \mid \gamma(w) = A\}. \quad (8)$$

That is, $j_F(A)$ consists of all those $w$'s which are compatible with every element in $A$ and not compatible with any element outside $A$.

**Theorem 1.** (Wong, Wang and Yao, 1991) Let $\underline{F}(A)$ and $\overline{F}(A)$ be two mappings from $2^\Theta$ to $2^W$. There exists a basic set assignment, $j_F : 2^\Theta \longrightarrow 2^W$, if and only if $F = (\underline{F}, \overline{F})$ is an interval structure.

At this point, one can clearly see the similarity between interval structures and belief/plausibility functions (Shafer, 1976), and the similarity between basic set assignments and basic probability assignments. Interval structures can be viewed as the non-numeric counterparts of belief/plausibility functions.

## 3 REPRESENTATIONS OF UNCERTAINTY BY INTERVAL STRUCTURES

The relationships between an interval structure and other representations of uncertainty will be explored in this section. We argue that the proposed interval structure provides a unified framework for these methods.

### 3.1 ROUGH SETS

In many applications, a *concept* may not be conveniently described by an ordinary (crisp) set. Pawlak (1982, 1984) introduced the notion of rough sets. With such a construct, a concept can be represented by a pair of ordinary sets referred to as the *lower* and *upper approximations*.

Let $\Theta$ denote the universe (a finite ordinary set), and let $R \subseteq \Theta \times \Theta$ be an equivalence (indiscernability) relation on $\Theta$, i.e., $R$ is reflexive, symmetric and transitive. The pair $Apr = (\Theta, R)$ is called an approximation space. The equivalence relation $R$ partitions the set $\Theta$ into disjoint subsets, denoted by $\Theta/R = \{w_1, w_2, \ldots, w_m\}$, where $w_i$ is an equivalence class of $R$. If two elements $\theta_1, \theta_2$ in $\Theta$ belong to the same equivalence class $w \in \Theta/R$, we say that $\theta_1$ and $\theta_2$ are indistinguishable. The equivalence classes of $R$ and the empty set $\emptyset$ are the elementary or atomic sets in the approximation space $Apr = (\Theta, R)$. The union of one or more elementary sets is called a composed set in $Apr$.

For an arbitrary concept $A \in 2^\Theta$, the lower and upper approximations are defined as:

$$\underline{Apr}(A) = \bigcup_{w_i \subseteq A} w_i, \quad \overline{Apr}(A) = \bigcup_{w_i \cap A \neq \emptyset} w_i. \quad (9)$$

That is, the lower approximation $\underline{Apr}(A)$ is the union of all the elementary sets which are subsets of $A$, and the upper approximation $\overline{Apr}(A)$ is the union of all the elementary sets which have a non-empty intersection with $A$. The interval $[\underline{Apr}(A), \overline{Apr}(A)]$ is the representation of an ordinary set $A$ in the approximation space $Apr = (\Theta, R)$, or simply called the rough set of $A$. By definition, $\underline{Apr}(A)$ is the greatest composed set contained in $A$, and $\overline{Apr}(A)$ is the least composed set containing $A$.

The notion of rough sets was also discussed in (Shafer, 1976; Wong and Lingras, 1989; Dubois and Prade, 1990). The quotient set $W = \Theta/R$ is a *coarsening* of $\Theta$, while $\Theta$ is a *refinement* of $W = \Theta/R$. In this special case, a multi-valued mapping, $\gamma : W \longrightarrow 2^\Theta$, can be defined as:

$$\gamma(w) = \{\theta \mid w = [\theta]_R\} = w, \quad (10)$$

where $[\theta]_R$ denotes the equivalence class to which $\theta$ belongs. Based on equations (3), (4) and (10), an interval can be defined as:

$$\underline{F}(A) = \{w | w \subseteq A\}, \quad \overline{F}(A) = \{w | w \cap A \neq \emptyset\}. \quad (11)$$

The sets $\underline{F}(A)$ and $\overline{F}(A)$ are called the inner and outer reductions, respectively, by Shafer (1976). Clearly, the lower and upper approximations of the rough-set model can be expressed in terms of the inner and outer reductions as follows:

$$\underline{Apr}(A) = \bigcup_{w \in \underline{F}(A)} w, \quad \overline{Apr}(A) = \bigcup_{w \in \overline{F}(A)} w. \quad (12)$$

Therefore, these constructs of rough sets can be interpreted as an interval structure. The properties of lower and upper approximations given by Pawlak (1982) immediately follow from the properties satisfied by an interval structure.

The rough-set model has been used successfully in pattern classification and for generating decision rules (Pawlak, 1984; Pawlak, Wong and Ziarko, 1988). For example, consider a medical diagnosis problem (Pawlak, Slowinski and Slowinski, 1986). Suppose $W$ is a set of symptoms, and $\Theta$ is a set of diseases. By the symptoms, one can divide the patients into subgroups. An element $w \in W$ is the *description* or *label* of a subgroup of patients with the same symptoms. Let $A$, a subset of $\Theta$, denote a set of diseases. In order to decide if a patient has contracted any of the diseases in $A$, the rough-set model suggests two kinds of decision rules:

$$\underline{F}(A) \rightarrow A, \qquad \overline{F}(A) \rightsquigarrow A, \quad (13)$$

where $F = (\underline{F}, \overline{F})$ is the interval structure defined by equation (11). The deterministic rule $\underline{F}(A) \rightarrow A$ indicates that if the patient has the symptoms in $\underline{F}(A)$, then he/she has definitely contracted the diseases in $A$. On the other hand, the non-deterministic decision rule $\overline{F}(A) \rightsquigarrow A$ indicates that a patient with symptoms in $\overline{F}(A)$ could suffer from the diseases in $A$. These deterministic and non-deterministic rules are governed by the properties of the interval structure.

The rough-set model outlined above considers a special kind of relationship between two sets, i.e., one set is a coarsening of the other. There are a number of extensions of the rough-set model. For example, instead of using an equivalence relation the rough-set model may be formulated by using a *compatibility* relation (i.e., reflexive and symmetric but not necessarily transitive) on $\Theta$. Dubois and Prade (1990) considered fuzzy similarity relations and fuzzy partitions for the approximation of sets, which lead to the notion of fuzzy rough sets.

## 3.2 INCIDENCE CALCULUS

In order to overcome the problems associated with using numeric methods for probabilistic reasoning, Bundy (1985, 1986) introduced incidence calculus, a technique for assigning uncertainty values to propositions. These uncertainty values are in fact sets of points called *incidences* which can be interpreted as classes of situations or possible worlds. The uncertainty of a proposition is its incidence.

Following Shafer (1976, 1986), for any question we can define a set $\Theta$ of all possible answers based on our knowledge, and we know that exactly one of these answers is correct. This set $\Theta$ is called a frame of discernment, or simply a frame. Any subset $A \subseteq \Theta$ is regarded as a proposition that the true answer lies in $A$. The power set $2^\Theta$ represents all possible propositions discerned by the frame $\Theta$. Such correspondence between propositions and subsets is useful because it translates the logical notions of conjunction, disjunction, implication, and negation into the more familiar set theoretic notions of intersection, union, inclusion, and complementation. We will use this representation of propositions in the discussion of incidence calculus.

Given a frame $\Theta$ and a set of incidences $W$, one can define a mapping $i : 2^\Theta \longrightarrow 2^W$. For any proposition $A \in 2^\Theta$, $i(A)$ is referred to as the incidence of $A$. The mapping $i : 2^\Theta \longrightarrow 2^W$ obeys the following axioms:

(I1) $\quad i(A \cup B) = i(A) \cup i(B)$,

(I2) $\quad i(\neg A) = W - i(A)$.

A mapping $i : 2^\Theta \longrightarrow 2^W$ satisfying axioms (I1) and (I2) is called an *incidence structure*. In practice, one may find it is difficult to specify precisely the incidence for each proposition. Instead, one may be able to provide the lower and upper *assignments* for the individual propositions. In other words, one can define two mappings $\inf : 2^\Theta \longrightarrow 2^W$ and $\sup : 2^\Theta \longrightarrow 2^W$ to indicate the interval within which the true incidence lies. The lower and upper assignments of incidences are consistent if there exists an incidence structure $i$ such that for every $A \in 2^\Theta$,

$$\inf(A) \subseteq i(A) \subseteq \sup(A); \tag{14}$$

$i$ is said to be bounded by the pair (inf, sup). A pair of assignments $\inf_0 : 2^\Theta \longrightarrow 2^W$ and $\sup_0 : 2^\Theta \longrightarrow 2^W$ represent the *tightest* bounds if (a) the pair $(\inf_0, \sup_0)$ is bounded by (inf, sup), (b) every incidence structure bounded by (inf, sup) is also bounded by $(\inf_0, \sup_0)$, namely, for all $A$:

$$\inf(A) \subseteq \inf_0(A) \subseteq i(A) \subseteq \sup_0(A) \subseteq \sup(A), \tag{15}$$

and (c) no other assignments within $(\inf_0, \sup_0)$ would satisfy conditions (a) and (b).

Since inf and sup are defined separately, these mappings are not necessarily consistent with each other. Bundy (1985, 1986) proposed a set of inference rules to test the consistency of the lower and upper assignments. If the assignments are consistent, the application of the inference rules will produce the tightest bounds for the individual propositions. We will demonstrate in Section 4 that these tightest bounds indeed satisfy the axioms of an interval structure.

## 3.3 BELIEF FUNCTIONS

We have shown that both the rough-set model and incidence calculus use an interval structure to represent non-numeric uncertain information. Now we want to show that the belief and plausibility functions, originating from the concepts of lower and upper probabilities induced by a multi-valued mapping (Dempster, 1967), can also be considered as an interval structure representing numeric uncertain information.

A belief function $Bel$ is a mapping from $2^\Theta$ to the interval [0, 1], $Bel : 2^\Theta \rightarrow [0, 1]$, satisfying the following axioms (Shafer, 1976; Dubois and Prade, 1986; Smets, 1988, 1990):

(B1) $\quad Bel(\emptyset) = 0$,

(B2) $\quad Bel(\Theta) = 1$,

(B3) $\quad$ For every positive integer $n$ and every collection $A_1, A_2, \ldots, A_n \in 2^\Theta$,

$$Bel(A_1 \cup A_2 \ldots \cup A_n) \geq \sum_i Bel(A_i) - \sum_{i<j} Bel(A_i \cap A_j) \pm \ldots (-1)^{n+1} Bel(A_1 \cap A_2 \ldots \cap A_n).$$

A belief function can be equivalently defined by another mapping, $m : 2^\Theta \rightarrow [0, 1]$, which is called a basic probability assignment satisfying:

(M1) $\quad m(\emptyset) = 0$,

(M2) $\quad \sum_{A \in 2^\Theta} m(A) = 1$.

In terms of the basic probability assignment, the belief in a subset $A \subseteq \Theta$ can be expressed as:

$$\text{(M3)} \qquad Bel(A) = \sum_{B \subseteq A} m(B).$$

A subset $A \in 2^{\Theta}$ with $m(A) > 0$ is called a *focal element*. By the Möbius inversion one can construct the basic probability assignment from a belief function (Shafer, 1976):

$$m(A) = \sum_{B \subseteq A} (-1)^{|A-B|} Bel(B), \qquad (16)$$

where $|\cdot|$ denotes the cardinality of a set. Therefore, a belief function can be defined by axioms (B1)-(B3) or (M1)-(M3).

For a given belief function, one can define another function called ***plausibility*** as follows:

$$Pl(A) = 1 - Bel(\neg A).$$

A plausibility function can be independently defined by the dual axiom of (B3). The belief in a subset $A \subseteq \Theta$ is interpreted as the belief one actually commits to $A$, whereas the plausibility of $A$ is interpreted as the maximum possible belief one may commit to $A$. It can be easily verified that $Pl(A) \geq Bel(A)$. The interval $[Bel(A), Pl(A)]$ represents the quantitative judgments on a proposition $A$ based on a given evidence.

The following theorems demonstrate the close relationships between belief functions and interval structures.

**Theorem 2.** (Wong, Wang and Yao, 1991) Let $W$ and $\Theta$ be two finite sets. Let $F = (\underline{F}, \overline{F})$ be an ***interval structure*** with $\underline{F} : 2^{\Theta} \longrightarrow 2^W$ and $\overline{F} : 2^{\Theta} \longrightarrow 2^W$. Suppose $P$ is a probability function on $W$. Then $P(\underline{F}(A))$ is a belief function and $P(\overline{F}(A))$ is the corresponding plausibility function.

**Theorem 3.** The mappings $Bel$ and $Pl$ from $2^{\Theta}$ to $[0, 1]$ are belief and plausibility functions, if and only if there exists an interval structure $F$ on a finite set $W$, and a probability $P$ on $W$ such that:

$$Bel(A) = P(\underline{F}(A)), \qquad Pl(A) = P(\overline{F}(A)). \qquad (17)$$

The ***if*** part of this theorem is essentially given by Theorem 2. The ***only if*** part of the theorem can be proved as follows. Suppose $Bel : 2^{\Theta} \rightarrow [0, 1]$ is a belief function. There exists a basic probability assignment $m : 2^{\Theta} \rightarrow [0, 1]$ such that $Bel(A) = \sum_{B \subseteq A} m(A)$. Each element $A$ with $m(A) \neq \emptyset$ is called a *focal element*. Based on the focal elements, one can construct a finite set $W$ as:

$$W = \{w_A \mid m(A) \neq 0\}.$$

The probability $P$ on $W$ may be defined as:

$$P(\{w_A\}) = m(A).$$

Based on the basic probability assignment $m$, one may define a basic mapping $j_F : 2^{\Theta} \rightarrow 2^W$ as:

$$j_F(A) = \begin{cases} \{w_A\} & \text{if } m(A) \neq 0 \\ \emptyset & \text{if } m(A) = 0. \end{cases}$$

Let $\underline{F}(A) = \bigcup_{B \subseteq A} j_F(B)$ and $\overline{F}(A) = W - \underline{F}(\neg A)$. By Theorem 1, $F = (\underline{F}, \overline{F})$ is an interval structure. Moreover,

$$\begin{aligned} P(\underline{F}(A)) &= \sum_{B \subseteq A} P(j_F(B)) \\ &= \sum_{B \subseteq A} P(\{w_B\}) \\ &= \sum_{B \subseteq A} m(B) = Bel(A), \end{aligned}$$

and

$$\begin{aligned} P(\overline{F}(A)) &= P(W - \underline{F}(\neg A)) \\ &= 1 - P(\underline{F}(\neg A)) \\ &= 1 - Bel(\neg A) = Pl(A). \end{aligned}$$

From the results of Theorems 2 and 3, it can be seen that belief/plausibility functions can be understood in terms of an interval structure. Clearly, the numeric axioms (B1)-(B3) correspond to the non-numeric axioms (L1)-(L4). The non-numeric and numeric bounds are connected by a probability function. Similar observations were also noted by Bundy (1985), Corred de Silva and Bundy (1990) in the study of incidence calculus.

In the above discussion, we have demonstrated that the rough-set model, incidence calculus, and belief/plausibility functions are all linked to an interval structure. Our analysis suggests that interval structures provide a common framework for representing uncertain information. Similarly, the different levels of approximations considered by Yao and Wong (1991) and the notion of fuzzy rough sets introduced by Dubois and Prade (1990) can also be interpreted as an interval structure.

# 4 KNOWLEDGE SYNTHESIS USING INTERVAL STRUCTURE

In the design of expert systems, decision rules can be directly given by the experts. There are two potential problems associated with such input knowledge. First, since these rules are specified separately for the individual propositions, inconsistency may occur. That is, there may exist contradictions among the given rules. Consider again the medical diagnosis problem. Suppose we have two rules, $r_1 : \{w_1, w_2\} \rightarrow \{\theta_1\}$ and $r_2 : \{w_1, w_2\} \rightsquigarrow \{\theta_2\}$. The first rule $r_1$ says that if symptom is $w_2$, disease is $\theta_1$, and the second rule $r_2$ implies that if symptom is $w_2$, disease is not $\theta_1$. Clearly, there exists a contradiction between the two rules $r_1$ and $r_2$. It is therefore necessary to test the consistency of the input rules. Secondly, new decision rules can be logically inferred from the given rules. For instance, from $\{w_1\} \rightarrow \{\theta_1, \theta_2\}$, we know that if symptom is $w_1$, disease is $\theta_1$ or $\theta_2$. Also, from another rule $\{w_1\} \rightarrow \{\theta_1, \theta_3\}$, we can conclude that if symptom is $w_1$, disease is $\theta_1$ or $\theta_3$. These two rules together imply a new decision rule, namely, $\{w_1\} \rightarrow \{\theta_1\}$. Thus, a method for synthesizing or consolidating such input knowledge is required.

For any $A \in 2^\Theta$, the experts can specify a subset $\underline{G}(A) \subseteq W$ as the *lower assignment* and a subset $\overline{G}(A) \subseteq W$ as the *upper assignment* of $A$. The lower and upper assignments define the right hand side of the deterministic and non-deterministic rules $\underline{G}(A) \rightarrow A$ and $\overline{G}(A) \rightsquigarrow A$, respectively. To be consistent with such interpretations, we may assume that $\underline{G}(A) \subseteq \overline{G}(A)$. Furthermore, if $\underline{G}(A)$ or $\overline{G}(A)$ is not given, we assume $\underline{G}(A) = \emptyset$ or $\overline{G}(A) = W$. The lower and upper assignments can be viewed as a pair of mappings $\underline{G}$ and $\overline{G}$ from $2^\Theta$ to $2^W$.

An interval structure $F = (\underline{F}, \overline{F})$ is *inside* a pair of lower and upper assignments $\underline{G}$ and $\overline{G}$ if for every $A \in 2^\Theta$,

$$\underline{G}(A) \subseteq \underline{F}(A) \subseteq \overline{F}(A) \subseteq \overline{G}(A). \tag{18}$$

Let $D_G$ be the set of decision rules associated with $G = (\underline{G}, \overline{G})$. We say that the rules in $D_G$ *logically* imply a deterministic rule $X \rightarrow A$, written $D_G \models X \rightarrow A$, if for every interval structure $F$ inside $G$, $X \subseteq \underline{F}(A)$ holds. Similarly, the rules in $D_G$ *logically* imply a non-deterministic rule $Y \rightsquigarrow A$, written $D_G \models Y \rightsquigarrow A$, if for every interval structure $F$ inside $G$, $\overline{F}(A) \subseteq Y$ holds. We use $D_G^*$ to denote the set of all rules that are logically implied by $D_G$.

Wong, Wang and Yao (1991) introduced the following set of inference axioms to derive $D_G^*$:

$(I_1)$ $X \rightsquigarrow A$ and $Y \rightarrow \neg A \Longrightarrow X - Y \rightsquigarrow A$.

$(I_2)$ $X \rightsquigarrow \neg A$ and $Y \rightarrow A \Longrightarrow Y \cup (W - X) \rightarrow A$.

$(I_3)$ $X \rightsquigarrow A$, $Y \rightsquigarrow B$ and $Z \rightsquigarrow A \cap B \Longrightarrow$ $X \cap Y \cap Z \rightsquigarrow A \cap B$.

$(I_4)$ $X \rightarrow A$, $Y \rightarrow B$ and $Z \rightarrow A \cap B \Longrightarrow$ $(X \cap Y) \cup Z \rightarrow A \cap B$.

$(I_5)$ $X \rightarrow A \cap B$ and $Y \rightarrow A \Longrightarrow X \cup Y \rightarrow A$.

$(I_6)$ $X \rightarrow A \Longrightarrow Y \rightarrow A$ for any $Y \subseteq X$.

$(I_7)$ $X \rightsquigarrow A \Longrightarrow Y \rightsquigarrow A$ for any $Y \supseteq X$.

Let $I$ denote a set of inference axioms. With respect to $I$, the *closure* of $D_G$, written $D_G^+$, is the *smallest* set containing $D_G$ such that the inference axioms cannot be applied to the set to yield a decision rule not in the set. The set of inference axioms is sound and complete if $D_G \models D_G^+$, i.e., any rule in $D_G^+$ is in $D_G^*$, and $D_G^* \subseteq D_G^+$. It has been shown by Wong, Wang and Yao (1991) that the above inference axioms $(I_1)$-$(I_7)$ are indeed both sound and complete.

For any $A$ in $2^\Theta$, $\underline{F}(A)$ is called the *max* lower bound of $A$, if for any $X \rightarrow A$ in $D_G^+$, $X \subseteq \underline{F}(A)$; $\overline{F}(A)$ is called the *min* upper bound of $A$, if for any $Y \rightsquigarrow A$ in $D_G^+$, $\overline{F}(A) \subseteq Y$. The max-min bounds are in fact the tightest bounds in the incidence calculus. As shown by the following theorem, these bounds satisfy the axioms of an interval structure.

**Theorem 4.** (Wong, Wang and Yao, 1991) The max-min bounds derived from a consistent assignment $G$ form an interval structure.

Recall that an interval structure can be equivalently defined by a basic set assignment. The results of Theorem 4 thus provide an alternative way to construct the max-min bounds. That is, one can construct the basic set assignment $j_F$ instead. The algorithm suggested by Wong, Wang and Yao (1991) for constructing the basic set assignment is outlined below.

**Input**: $G = \{\underline{G}(A) \rightarrow A,\ \overline{G}(A) \rightsquigarrow A \mid$
$A \in 2^\Theta,\ \underline{G}(A) \neq \emptyset$ and $\overline{G}(A) \neq W\}$;

1. **for** each rule $\overline{G}(A) \rightsquigarrow A$ in $G$ **do**
   $\underline{G}'(\neg A) = \underline{G}(\neg A) \cup (W - \overline{G}(A))$;
2. **for** each $w_k \in W$ **do**
   Find all the $A$'s where $\underline{G}'(A) \neq \emptyset$ such that $w_k \in \underline{G}'(A)$, say, $A_1, A_2, \ldots, A_l$;
   **if** $A_1 \cap A_2 \cap \ldots \cap A_l = \emptyset$ **then**
   exits to *no interval structure*;
   else
   $j(A_1 \cap A_2 \cap \ldots \cap A_l) = j(A_1 \cap A_2 \cap \ldots \cap A_l) \cup \{w_k\}$;
   (Initially, $j(A_1 \cap A_2 \cap \ldots \cap A_l) = \emptyset$.)
3. **Output**: $j$.

In step 1 of the above procedure, if $\underline{G}(\neg A)$ is not assigned a value in the input, we assume the value is $\emptyset$. Moreover, if the input value $\underline{G}(A)$ is not changed, we also denote it by $\underline{G}'(A)$. It is understood that all those initial assignments with $\underline{G}(A) = \emptyset$ and $\overline{G}(A) = W$ have been eliminated from the input.

The following example illustrates the proposed procedure for constructing the basic set assignment and the max-min bounds.

**Example.** Let $W = \{w_1, w_2, w_3, w_4, w_5\}$ and $\Theta = \{\theta_1, \theta_2, \theta_3\}$. Suppose the initial lower and upper assignments are given as:

$$\underline{G}(\{\theta_1, \theta_2\}) = \{w_1, w_4\},$$

$$\underline{G}(\{\theta_1,\ \theta_3\}) = \{w_1,\ w_2\},$$
$$\underline{G}(\Theta) = \{w_3\},$$
$$\overline{G}(\{\theta_3\}) = \{w_3,\ w_5\},$$
$$\overline{G}(\{\theta_1\}) = \{w_1,\ w_2,\ w_3\}.$$

In step 1, the given two upper assignments yield:

$$\begin{aligned}\underline{G}'(\neg\{\theta_3\}) &= \underline{G}'(\{\theta_1,\ \theta_2\})\\ &= \underline{G}(\{\theta_1,\theta_2\}) \cup (W - \overline{G}(\{\theta_3\})\\ &= \{w_1, w_4\} \cup (W - \{w_3, w_5\})\\ &= \{w_1,\ w_2,\ w_4\},\end{aligned}$$

$$\begin{aligned}\underline{G}'(\neg\{\theta_1\}) &= \underline{G}'(\{\theta_2,\ \theta_3\})\\ &= \underline{G}(\{\theta_2,\theta_3\}) \cup (W - \overline{G}(\{\theta_1\})\\ &= \emptyset \cup (W - \{w_1, w_2, w_3\})\\ &= \{w_4,\ w_5\}.\end{aligned}$$

Thus, together with the given lower assignments, we obtain:

$$\underline{G}'(\{\theta_1,\ \theta_2\}) = \{w_1,\ w_2,\ w_4\},$$
$$\underline{G}'(\{\theta_1,\ \theta_3\}) = \{w_1,\ w_2\},$$
$$\underline{G}'(\{\theta_2,\ \theta_3\}) = \{w_4,\ w_5\},$$
$$\underline{G}'(\Theta) = \{w_3\}.$$

In step 2, since

$$w_1 \in \underline{G}'(\{\theta_1,\ \theta_2\}), \qquad w_1 \in \underline{G}'(\{\theta_1,\ \theta_3\}),$$

it follows:

$$w_1 \in j(\{\theta_1,\ \theta_2\} \cap \{\theta_1,\ \theta_3\}) = j(\{\theta_1\}).$$

Similarly,

$$w_2 \in j(\{\theta_1\}),$$
$$w_3 \in j(\Theta),$$
$$w_4 \in j(\{\theta_2\}),$$
$$w_5 \in j(\{\theta_2,\ \theta_3\}).$$

Therefore, the basic set assignment $j_F$ is given by:

$$j_F(\{\theta_1\}) = \{w_1,\ w_2\},$$
$$j_F(\{\theta_2\}) = \{w_4\},$$
$$j_F(\{\theta_2,\ \theta_3\}) = \{w_5\},$$
$$j_F(\Theta) = \{w_3\}.$$

By using the formulas:

$$\underline{F}(A) = \bigcup_{B \subseteq A} j_F(B)$$

and

$$\overline{F}(A) = \bigcup_{A \cap B \neq \emptyset} j_F(B),$$

one can easily construct the max lower bounds and the min upper bounds for every $A \in 2^{\Theta}$:

$$\underline{F}(\emptyset) = \emptyset,$$
$$\underline{F}(\{\theta_1\}) = \{w_1,\ w_2\},$$
$$\underline{F}(\{\theta_2\}) = \{w_4\},$$
$$\underline{F}(\{\theta_3\}) = \emptyset,$$
$$\underline{F}(\{\theta_1,\ \theta_2\}) = \{w_1,\ w_2,\ w_4\},$$
$$\underline{F}(\{\theta_1,\ \theta_3\}) = \{w_1,\ w_2\},$$
$$\underline{F}(\{\theta_2,\ \theta_3\}) = \{w_4, w_5\},$$
$$\underline{F}(\Theta) = W,$$

and

$$\overline{F}(\emptyset) = \emptyset,$$
$$\overline{F}(\{\theta_1\}) = \{w_1,\ w_2,\ w_3\},$$
$$\overline{F}(\{\theta_2\}) = \{w_3,\ w_4,\ w_5\},$$
$$\overline{F}(\{\theta_3\}) = \{w_3,\ w_5\},$$
$$\overline{F}(\{\theta_1,\ \theta_2\}) = W,$$
$$\overline{F}(\{\theta_1,\ \theta_3\}) = \{w_1,\ w_2,\ w_3,\ w_5\},$$
$$\overline{F}(\{\theta_2,\ \theta_3\}) = \{w_3,\ w_4,\ w_5\},$$
$$\overline{F}(\Theta) = W.$$

This example clearly demonstrates that the proposed algorithm for finding the basic set assignment is more efficient than that of finding the tightest bounds directly (Bundy, 1985, 1986; Wong, Wang and Yao, 1991).

## 5 CONCLUSION

To make decisions under uncertainty, it is crucial to choose an appropriate structure to represent the uncertain information. In this paper, we have introduced a unified framework for representing uncertainty based on the notion of an interval structure. In this approach, lower and upper bounds are used to characterize a concept or an incidence. It is also shown that an interval structure can be equivalently defined by a basic set assignment. An interval structure may be considered as the non-numeric counterpart of belief and plausibility functions, while the basic set assignment as the non-numeric counterpart of the basic probability assignment.

With the proposed framework, we have demonstrated that the lower and upper approximations of the rough-set model, the lower and upper bounds in incidence calculus, and the belief and plausibility functions all obey the axioms of an interval structure. We believe that the notion of an interval structure greatly facilitates the study of the various representations of uncertainty.

An interval structure can be used to synthesize the decision rules provided by experts. We have introduced a set of both sound and complete inference axioms to perform such a task, and developed an efficient algorithm for finding the desirable set of decision rules.

## References

A. Bundy (1985). Incidence calculus: a mechanism for probabilistic reasoning. *Journal of Automated Reasoning,* **1**: 263-283.

A. Bundy (1986). Correctness criteria of some algorithms for uncertain reasoning using incidence calculus *Journal of Automated Reasoning,* **2**: 109-126.

F. Correa da Silva, A. Bundy (1990). On some equivalence relations between incidence calculus and Dempster-Shafer theory of evidence. *Proceedings of 6th International Workshop on Uncertainty in Artificial Intelligence*, 378-383, Cambridge, MA.

A.P. Dempster (1967). Upper and lower probabilities induced by a multivalued mapping. *Annals of Mathematical Statistics*, **38**: 325-339.

D. Dubois, H. Prade (1985) Evidence measure based on fuzzy information. *Automatica*, **21**: 547-562.

D. Dubois, H. Prade (1986). A set-theoretic view of belief functions: logical operations and approximations by fuzzy sets. *International Journal of General Systems*, **12**: 193-226.

D. Dubois, H. Prade (1990). Rough fuzzy sets and fuzzy rough sets. *International Journal of General Systems*, **17**: 191-209.

J. Grzymala-Busse (1987). Rough-set and Dempster-Shafer approaches to knowledge acquisition under uncertainty - a comparison. Department of Computer Science, University of Kansas (Manuscript).

J.Y. Halpern, R. Fagin (1990). Two views of belief: belief as generalized probability and belief as evidence. Proceedings of National Conference on Artificial Intelligence (AAAI-90), 112-119. An expended version appears as IBM Research Report RJ 7221, June, 1990.

Z. Pawlak (1982). Rough sets. *International Journal of Computer and Information and Sciences*, **11**: 341-356.

Z. Pawlak (1984). Rough classification. *International Journal of Man-Machine Studies*, **20**: 469-483.

Z. Pawlak, K. Slowinski, R. Slowinski (1986). Rough classification of patients after highly selective vagotomy for duodendal ulcer. *International Journal of Man-Machine Studies*, **24**: 413-433.

Z. Pawlak, S.K.M. Wong, W. Ziarko (1988). Rough sets: probabilistic versus deterministic approach. *International Journal of Man-Machine Studies*, **29**: 81-95.

G. Shafer (1976). *A Mathematical Theory of Evidence*, Princeton: Princeton University Press.

G. Shafer (1986). Belief functions and possibility measures. in *Analysis of Fuzzy Information*, Vol. I, J.C. Bezdek, Ed., CRC Press, 51-84.

P. Smets (1988). Belief functions (with discussions). *Non-standard Logics for Automated Reasoning*, P. Smets, A. Mamdani, D. Dubois, H. Prade, Eds., New York: Academic Press, 253-285.

P. Smets (1990). The combination of evidence in the transferable belief model. *IEEE Transaction on Pattern Analysis and Machine Intelligence*, **12**: 447-458.

S.K.M. Wong, P. Lingras (1989). The compatibility view of Shafer-Dempster theory using the concept of rough set. *Methodologies for Intelligent Systems, 4*. Z.W. Ras, Ed., New York: North-Holland, 33-42.

S.K.M. Wong, L.S. Wang, Y.Y. Yao (1991). Knowledge synthesis in rule-based systems. Submitted for publication.

Y.Y. Yao, S.K.M. Wong (1991). A decision theoretic framework for approximating concepts. To appear in *International Journal of Man-machine Studies*.

L.A. Zadeh (1965). Fuzzy sets. *Information and Control*, **8**: 338-353.